\documentclass[10pt,twocolumn,twoside]{IEEEtran}
\usepackage{amsmath}
\usepackage[utf8]{inputenc}
\usepackage{bm}
\usepackage{amsfonts}
\usepackage{graphicx}
\usepackage{graphics}
\usepackage{amssymb,amsthm}
\usepackage{algorithmicx}
\usepackage{algorithm}
\usepackage{algpseudocode}
\usepackage{multirow}
\usepackage{xcolor}
\usepackage{url}
\usepackage[stable]{footmisc}

\newcommand{\be}{\begin{equation}}
	\newcommand{\ee}{\end{equation}}
\newcommand{\ba}{\begin{array}}
	\newcommand{\ea}{\end{array}}
\newcommand{\baa}{\left[\begin{array}}
	\newcommand{\eaa}{\end{array}\right]}

\newcommand{\beqa}{\begin{eqnarray}}
	\newcommand{\eeqa}{\end{eqnarray}}
\newcommand{\bt}{\begin{tabular}}
	\newcommand{\et}{\end{tabular}}

\newcommand{\bi}{\begin{itemize}}
	\newcommand{\ei}{\end{itemize}}
\newcommand{\bc}{\begin{center}}
	\newcommand{\ec}{\end{center}}

\title{An iterative coordinate descent algorithm to compute sparse low-rank approximations}

\author{Cristian Rusu%, \IEEEmembership{Member, IEEE}
	\thanks{C. Rusu is with the
		Faculty of Automatic Control and Computer Science,
		University Politehnica of Bucharest, Romania
		(email: cristian.rusu@upb.ro) and with the Research Center for Logic, Optimization and Security (LOS),
		Department of Computer Science,
		Faculty of Mathematics and Computer Science,
		University of Bucharest,
		Romania. This work was supported by
		the Romanian Ministry of Education and Research, CNCS-UEFISCDI,
		project number PN-III-P1-1.1-TE-2019-1843, within PNCDI III.
}}

\begin{document}
	\maketitle
	\begin{abstract}
		In this paper, we describe a new algorithm to build a few sparse principal components from a given data matrix. Our approach does not explicitly create the covariance matrix of the data and can be viewed as an extension of the Kogbetliantz algorithm to build an approximate singular value decomposition for a few principal components. We show the performance of the proposed algorithm to recover sparse principal components on various datasets from the literature and perform dimensionality reduction for classification applications.
	\end{abstract}
	
	
	%---------------------------------------------------------------------
	\section{Introduction}
	%---------------------------------------------------------------------
	
	%{\color{red} meaningless intro}
	
	The singular value decomposition (SVD) is one of the cornerstone algorithms in numerical linear algebra with numerous applications to signal processing.
	
	%{\color{red} SVD is well understood}
	
	To perform the complete SVD we have available well-established algorithms (reduction to bidiagonal form via Householder reflectors followed by the QR algorithm \cite{trefethen97}[Lecture 31]). But, in many applications, we are interested to compute only a partial, approximate decomposition. For example, in dimensionality reduction (DR) applications such as principal component analysis (PCA) we are interested in computing only a few of the singular vectors associated with the highest singular values. 
	
	%{\color{red} a brief description of SVD and PCA, introduce notation}
	
	Given a data matrix $\mathbf{X} \in \mathbb{R}^{n \times N}$, where $n$ is the number of features and $N$ is the number of data points, the SVD computes the factorization $\mathbf{X} = \mathbf{U \Sigma V}^T$ where $\mathbf{U}$ and $\mathbf{V}$ are orthonormal matrices of appropriate sizes which contain the left and right singular vectors, respectively, and the diagonal matrix $\mathbf{\Sigma}$ contains the positive singular values $\sigma_k$ in decreasing order. The first $p$ columns of $\mathbf{U}$, denoted $\mathbf{U}_p$, represent the first $p$ principal components from the data. We have that $\mathbf{U}_p$ maximizes $\| \mathbf{U}_p^T \mathbf{X} \|_F$ with the orthogonality constraint that $\mathbf{U}_p^T \mathbf{U}_p = \mathbf{I}_{p},\ p \leq n$. PCA is a non-convex optimization problem but due to the SVD, we are able to solve it exactly.
	
	%{\color{red} sparse singular vectors are also important}
	
	For interpretability and consistency issues \cite{10.2307/40592215}, researchers have introduced the sparse principal component analysis (sPCA): PCA with the additional constraint that the singular vectors to compute are sparse, i.e., $\mathbf{U}_p$ is now sparse and orthonormal. This approach has seen many signal processing applications such as image decomposition and texture segmentation \cite{4639574}, shape encoding in images \cite{8066311}, compressed hyperspectral imaging \cite{7219458}, target localization in hyperspectral imaging \cite{9019651}, and moving object detection in videos \cite{8485415}.
	
	%{\color{red} literature review for sparse singular vectors}
	
	The extra sparsity constraint for sPCA means that the straightforward SVD can no longer provide the optimal solution and therefore the non-convex optimization problem is hard to solve in general (orthogonality \cite{984753} and sparsity constraints \cite{6873279}). For this reason, many different approaches have been proposed in the literature to approximately solve the problem. sPCA was first introduced in \cite{doi:10.1080/757584614} using simple techniques such as canceling the lowest absolute values entries in the singular vectors (losing orthogonality in the process). Following this work, several LASSO and convex optimization (particularly semidefinite programming) approaches were developed to deal with the sparsity of the singular vectors \cite{doi:10.1198/1061860032148, 10.2307/2346178, doi:10.1198/106186006X113430, 10.2307/20453990}. There are numerous ways to deal with the sPCA and these include approaches such as: greedy methods \cite{10.5555/1390681.1442775}, geodesic steepest descent \cite{4637857}, Givens rotations \cite{4479581}, low rank approximations via a regularized (sparsity promoting) singular value decomposition \cite{SHEN20081015}, truncated power iterations \cite{Hein2010AnIP, 10.5555/1756006.1756021, 10.2307/23566581}, steepest descent on the Stiefel manifold using rotations matrices \cite{7852242} or on the Grassmannian manifold \cite{8574941}, quasi-Newton optimization for the sparse generalized eigenvalue problem \cite{8450790}, iterative deflation techniques \cite{9022486}, the minorization-maximization framework \cite{Song2015SparseGE} with additional orthogonality constraints \cite{7558183} or on the Stiefel manifold \cite{9354027}, 
	
	%{\color{red} what we propose}
	
	We propose a new way to build the left singular vectors by operating directly on a given data matrix (without constructing its covariance matrix) using a product of basic orthonormal transformations that work on two coordinates at a time. Orthogonality is maintained at every step of the algorithm and sparsity is controlled by the number of basic transformations used. The method can be seen as an extension of the Kogbetliantz algorithm \cite{10.2307/2098862, 7082836, 7096746} for the case when only a subset $p$ of the most important (sparse) principal components are to be computed. We are not concerned with the choice of $p$ \cite{6851153}, we assume $p$ given and fixed.
	
	%{\color{red} paper structure}
	The paper is structured as follows: Section II described the proposed method highlighting the three main ingredients of the procedure, Section III gives experimental support for the proposed method and Section IV concludes the paper.
	
	%---------------------------------------------------------------------
	\section{The proposed method}
	%---------------------------------------------------------------------
	
	In this section, we detail the proposed method to approximately compute sparse singular vectors. We start by describing the constrained least-squares optimization problem that we want to solve and then proceed to add structured matrices such that closed-form, locally optimal, updates can be iteratively applied to approximately solve this optimization problem. 
	
	\subsection{The optimization problem}
	
	We begin by describing the constrained optimization problem whose solutions are the singular vectors of the data matrix.
	
	\noindent \textbf{Result 1.} Given a data matrix $\mathbf{X} \in \mathbb{R}^{n \times N}$, assuming $N \geq n$ or even $N \gg n$, i.e., data points are stored columnwise, and denoting $\mathbf{W} = \begin{bmatrix} \mathbf{I}_{p} & \mathbf{0}_{p \times (N-p)} \\ \mathbf{0}_{(n-p) \times p} & \mathbf{0}_{(n-p)\times (N-p)} \end{bmatrix} $ the solution to the following optimization problem
	\begin{equation}
		\begin{aligned}
			& \underset{ \mathbf{\bar{U}}, \mathbf{\bar{V}}}{\text{minimize}} && \|  \mathbf{W} - \mathbf{\bar{U}}^T \mathbf{X} \mathbf{\bar{V}}  \|_F^2 \\
			& \text{subject to} & & \mathbf{\bar{U}}^T\mathbf{\bar{U}} = \mathbf{I}_{n}, \mathbf{\bar{V}}^T\mathbf{\bar{V}} = \mathbf{I}_{ N },
		\end{aligned}
		\label{eq:optimizationproblem}
	\end{equation}
	is given when $\mathbf{\bar{U}} = \mathbf{U}$ and $\mathbf{\bar{V}} = \mathbf{V}$ from the SVD $\mathbf{X} = \mathbf{U \Sigma V}^T$. The minimum value is $\sum_{k=1}^p (\sigma_k - 1)^2 + \sum_{k=p+1}^n \sigma_k^2$.% where we have used the largest $p$ singular values $\sigma_k$.
	
	\noindent \textbf{Proof.} We have the SVD $\mathbf{X} = \mathbf{U \Sigma V}^T$. Use the invariance of the Frobenius norm to orthonormal transformations to reach:
	\begin{equation}
		\|  \mathbf{W} - \mathbf{\bar{U}}^T \mathbf{X} \mathbf{\bar{V}} \|_F^2 \! = \! \|  \mathbf{\bar{U}} \mathbf{W\bar{V}}^T - \mathbf{X}  \|_F^2 \! = \! \|  \mathbf{\bar{U}} \mathbf{W\bar{V}}^T - \mathbf{U \Sigma V}^T  \|_F^2.
		\label{eq:calculation}
	\end{equation}
	We have now reached the classic problem of finding the closest orthonormal matrix, the Procrustes problem \cite{Proc}, whose solution is given by $\mathbf{\bar{U}} = \mathbf{U}$ and $\mathbf{\bar{V}} = \mathbf{V}$. For this choice, the quantity in \eqref{eq:calculation} reduces to $\| \mathbf{U} \mathbf{W} \mathbf{V}^T -   \mathbf{U}  \mathbf{\Sigma} \mathbf{V}^T \|_F^2 =  \| \mathbf{W} - \mathbf{\Sigma} \|_F^2$.$\hfill \blacksquare$
	
	Solving this constrained optimization problem directly, without using the SVD, is computationally hard. We will next add an additional structural constraint to simplify the problem.
	
	\subsection{The proposed structure}
	
	We propose to factorize the singular vector matrix $\mathbf{\bar{U}}$ as a product of basic transformations that are easy to manipulate and for which we can write closed-form solutions to \eqref{eq:optimizationproblem}. 
	
	Consider the following $n \times n$ orthonormal linear transformation, called a G-transform \cite{FastSparsifyingTransforms, FastPCA}:
	\begin{equation}
		\mathbf{G}_{ij} = \begin{bmatrix} \mathbf{I}_{i-1} &  & &\\
			& * & & *  \\
			& & \mathbf{I}_{j-i-1} & \\
			& * &  & *  \\
			& & & & \mathbf{I}_{n-j} \\
		\end{bmatrix},
		\label{eq:theG}
	\end{equation}
	with $\mathbf{\tilde{G}} \in \left\{ \begin{bmatrix} c & -s \\ s & c \end{bmatrix}, \begin{bmatrix} c & s \\ s & -c	\end{bmatrix}  \right\}$, such that $c^2 + s^2 = 1$, where the non-zero part (denoted by $*$ and $\mathbf{\tilde{G}}$) is only on rows and columns $i$ and $j$. These transformations are generalization of Givens rotations that now also include reflection transformations. When optimizating with this structure, we can consider that we are doing coordinate descent with an orthogonality constraint. We need two coordinates $(i,j)$ at a time as on a single coordinate the orthonormal transformation would just be $\{\pm 1\}$ (the exact identity matrix or one where a single diagonal element was flipped to $-1$).
	
	We propose to write our unknowns $\mathbf{\bar{U}}$ and $\mathbf{\bar{V}}$ as a product of such transformations \eqref{eq:theG} which we denote by $\mathbf{G}_{ij}$ and $\mathbf{H}_{ij}$ in order to distinguish between the left and right singular vectors, respectively:
	\begin{equation}
		\begin{aligned}
			\mathbf{\bar{U}} = & \prod_{k=1}^m \mathbf{G}_{ i_k j_k } = \mathbf{G}_{ i_1 j_1 }  \cdots \mathbf{G}_{ i_m j_m }, \\
			\mathbf{\bar{V}} = & \prod_{k=1}^m \mathbf{H}_{ i_k j_k } = \mathbf{H}_{ i_1 j_1 }  \cdots \mathbf{H}_{ i_m j_m },
		\end{aligned}
		\label{eq:bothfactorizations}
	\end{equation}
	where $m \geq 1$ is fixed. Then we have that $\mathbf{\bar{U}}_p$ and $\mathbf{\bar{V}}_p$ consist of the first $p$ columns of $\mathbf{\bar{U}}$ and $\mathbf{\bar{V}}$, respectively.
	
	\subsection{The proposed solution}
	
	We now move to solve the optimization problem in \eqref{eq:optimizationproblem} by using the structure \eqref{eq:bothfactorizations} for the solutions, i.e., we minimize
	\begin{equation}
		\| \mathbf{W} - \mathbf{G}_{ i_m j_m }^T \cdots  \mathbf{G}_{ i_1 j_1 }^T \mathbf{X}   \mathbf{H}_{ i_1 j_1 }  \cdots   \mathbf{H}_{ i_m j_m } \|_F^2.
	\end{equation}
	
	Therefore, we assume that the matrices $\mathbf{\bar{U}}_p$ and $\mathbf{\bar{V}}_p$ are updated by adding an extra basic G-transform in its factorization. Assume that we have intialized the first $k-1$ G-transforms in \eqref{eq:theG} in order to reduce $\|  \mathbf{W} - \mathbf{G}_{i_{k-1} j_{k-1} }^T  \cdots \mathbf{G}_{i_{1} j_{1} }^T  \mathbf{X} \mathbf{H}_{i_{1} j_{1} }  \cdots \mathbf{H}_{i_{k-1} j_{k-1} }  \|_F^2$ then to optimally update the $k^\text{th}$ transform we have the following result. Herein, we denote $\mathbf{X}_k = \mathbf{G}_{i_{k-1} j_{k-1} }^T  \cdots \mathbf{G}_{i_{1} j_{1} }^T  \mathbf{X} \mathbf{H}_{i_{1} j_{1} }  \cdots \mathbf{H}_{i_{k-1} j_{k-1} }$.
	
	\noindent \textbf{Result 2.} Given a data matrix $\mathbf{X}_k \in \mathbb{R}^{n \times N}$ then we have
	\begin{equation}
		\text{min}\|  \mathbf{W} - \mathbf{G}_{i_k j_k}^T \mathbf{X}_k \mathbf{H}_{i_k j_k} \|_F^2\! \!=\! p+ \| \mathbf{X}_k \|_F^2 -2\text{tr}(\mathbf{W}^T\mathbf{X}_k) - 2\mathcal{C}_{i_k j_k}.
		\label{eq:themin}
	\end{equation}
	where $\mathcal{C}_{i_k j_k} = \max \{  \sqrt{ (X^{(k)}_{i_k i_k} + X^{(k)}_{j_k j_k})^2 +  (X^{(k)}_{i_k j_k} - X^{(k)}_{j_k i_k})^2 }, \\ \sqrt{  (X^{(k)}_{i_k i_k} - X^{(k)}_{j_k j_k})^2 +  (X^{(k)}_{i_k j_k} + X^{(k)}_{j_k i_k})^2  }  \} - X^{(k)}_{i_k i_k} - X^{(k)}_{j_k j_k}$ for the choices $\mathbf{G}_{i_k j_k}  = \mathbf{A}$ and $\mathbf{H}_{i_k j_k} = \mathbf{B}$ from the SVD of the $2\times 2$ matrix $\begin{bmatrix}  X^{(k)}_{i_k i_k} & X^{(k)}_{i_k j_k} \\ X^{(k)}_{j_k i_k} & X^{(k)}_{j_k j_k}  \end{bmatrix} = \mathbf{A S B}^T$.
	
	\noindent \textbf{Proof.} We follow and adapt the proof of Theorem 1 from \cite{FastPCA} and we develop the Frobenius norm quantity like:
	\begin{equation}
		\begin{aligned}
			\|  \mathbf{W} - & \mathbf{G}_{i_k j_k}^T \mathbf{X}_k \mathbf{H}_{i_k j_k}  \|_F^2 =  \| \mathbf{W} \|_F^2 + \|\mathbf{G}_{i_k j_k}^T \mathbf{X}_k \mathbf{H}_{i_k j_k}\|_F^2 \\
			& \quad \quad \quad \quad \quad \quad \quad \quad \quad - 2\text{tr}( \mathbf{W}^T\mathbf{G}_{i_k j_k}^T \mathbf{X}_k \mathbf{H}_{i_k j_k}   )  \\
			= & p + \| \mathbf{X}_k \|_F^2 - 2\text{tr}( \mathbf{W}^T\mathbf{G}_{i_k j_k}^T \mathbf{X}_k \mathbf{H}_{i_k j_k}   ).
		\end{aligned}
		\label{eq:quantitytominimize}
	\end{equation}
	We now focus on the trace term. Let $\mathbf{Z} = \mathbf{W}^T \mathbf{X}_k$ and let $\mathbf{\tilde{X}}_k =  \begin{bmatrix}
		X^{(k)}_{i_k i_k} & X^{(k)}_{i_k j_k} \\ X^{(k)}_{j_k i_k} & X^{(k)}_{j_k j_k}
	\end{bmatrix}$, a $2 \times 2$ submatrix of $\mathbf{X}_k$ on indices $i_k$ and $j_k$. Because $\mathbf{G}_{i_k j_k}$ and $\mathbf{H}_{i_k j_k}$ operate only on two coordinates we have that
	\begin{equation}
		\begin{aligned}
			\text{tr}(\mathbf{W}^T & \mathbf{G}_{i_k j_k}^T \mathbf{X}_k \mathbf{H}_{i_k j_k}  )  =   \text{tr}(\mathbf{Z}) + \text{tr}( \mathbf{\tilde{G}}^T   \mathbf{\tilde{X}}_k \mathbf{\tilde{H}} ) - \text{tr}( \mathbf{\tilde{X}}_k)     \\
			= & \sum_{q = 1, q \notin \{ i_k,j_k \} }^{p} Z_{q q} + \text{tr}(\mathbf{\tilde{G}}^T   \mathbf{\tilde{X}}_k \mathbf{\tilde{H}} ).
		\end{aligned}
		\label{eq:quantitytomaximize}
	\end{equation}
	In order to minimize the quantity in \eqref{eq:quantitytominimize} we need to maximize the trace term in \eqref{eq:quantitytomaximize} and the quantity $ \text{tr}( \mathbf{\tilde{G}}^T   \mathbf{\tilde{X}}_k \mathbf{\tilde{H}} )$. To maximize this, we use the two-sided Procrustes problem \cite{Schonemann1968}, the solution is given by the SVD of the $2 \times 2$ matrix $\mathbf{\tilde{X}}_k$. For this optimal choice, the objective function is given by $p + \| \mathbf{X}_k \|_F^2 - 2\text{tr}(\mathbf{Z}) - 2 ( \|  \mathbf{\tilde{X}}_k \|_* - \text{tr}(\mathbf{\tilde{X}}_k) )$ where the critical quantity is the difference between the sum of singular values (nuclear norm) and the sum of eigenvalues (trace) of $\mathbf{\tilde{X}}_k$. The quantities used for the calculation of $C_{i_k j_k}$ are reached by using the explicit formulas for the singular values of $\mathbf{\tilde{X}}_k$, i.e., $s_{1,2} = \sqrt{ \frac{1}{2} \left(  \| \mathbf{\tilde{X}}_k \|_F^2 \pm \sqrt{ \| \mathbf{\tilde{X}}_k\|_F^4  - 4 \det (\mathbf{\tilde{X}}_k)^2  }  \right) }$.$\hfill \blacksquare$
	
	Based on this result, we will choose to construct iteratively the factorizations in \eqref{eq:bothfactorizations} for each $k$ at a time such that at each step we locally minimize the objective function, i.e.,:
	\begin{equation}
		( i_k^\star, j_k^\star ) = \arg \max \mathcal{C}_{ i j }, i = 1,\dots, p, j = i+1, \dots, N.
		\label{eq:theC}
	\end{equation}

	As this is the main result of the paper, some clarifying remarks, and connections to previous work are in order.

	\noindent \textbf{Remark 1 (Choosing indices $(i_k, j_k)$).} The ranges on the indices are described in \eqref{eq:theC} but note that, because $\mathbf{X}_k$ is not square, we are out of bounds whenever $j_k > n$ because of the number of lines in $\mathbf{X}_k$. In this case, we take $\mathbf{\tilde{X}}_k =  \begin{bmatrix}
		X^{(k)}_{i_k i_k} & X^{(k)}_{i_k j_k} \\ 0 & 0
	\end{bmatrix}$ and therefore $\mathbf{G}_{i_k j_k} = \mathbf{I}_n$, i.e., there is no update to the left singural vectors but only on $\mathbf{\bar{V}}$. $\hfill \blacksquare$

	\noindent \textbf{Remark 2 (Connection to the Kogbetliantz method).} The proposed structures and solutions are based on iterative methods such as the Jacobi diagonalization method for symmetric matrices \cite{JacobiProcess} and the Kogbetliantz method for general matrices \cite{Kogbetliantz}. This is because we make use of updates $\mathbf{X}_{k+1} \leftarrow  \mathbf{G}_{i_k j_k}^T \mathbf{X}_k \mathbf{H}_{ i_k j_k } $ to diagonalize the data matrix $\mathbf{X}$. The significant difference with the proposed method is the way we choose indices $(i_k, j_k)$ to operate on: instead of using the classic largest off-diagonal absolute value from $\mathbf{X}_k$ we choose the indices and the orthonormal transformation on those indices that locally maximizes the trace of $\mathbf{X}_{k+1}$. The scores $\mathcal{C}_{ij}$ are always non-negative (nuclear norm is always larger than the trace for a given matrix \cite{Horn}) and zero only when $\mathbf{\tilde{X}}_k$ is positive definite. Note that $\mathbf{X}_{k+1}$ has zeroes in positions $(i_k, j_k)$ and $(j_k, i_k)$.$\hfill \blacksquare$
	
	\noindent \textbf{Remark 3 (Connection to previous work).} The matrix structure in \eqref{eq:theG} and factorizations like \eqref{eq:bothfactorizations} have been previously used to compute factorizations of orthonormal matrices where the goal was to maximize quantities such as $\text{tr}(\mathbf{Q}_{ i_k j_k } \mathbf{X}_k)$ with an orthonormal $\mathbf{Q}_{ i_k j_k }$, i.e., computing a polar decomposition of $\mathbf{X}_k$. In our optimization problem we use a two-sided transformation that maximizes $\text{tr}(\mathbf{G}_{i_k j_k}^T \mathbf{X}_k \mathbf{H}_{ i_k j_k })$, i.e., computing the singular value decomposition. We note that, assuming appropriate dimensions of matrices, because of the trace equality $\text{tr}(\mathbf{G}_{i_k j_k}^T \mathbf{X}_k \mathbf{H}_{ i_k j_k }) = \text{tr}(\mathbf{H}_{ i_k j_k } \mathbf{G}_{i_k j_k}^T \mathbf{X}_k)$ the two trace quantities are maximized for the same choices of indices when $\mathbf{Q}_{ i_k j_k }  = \mathbf{H}_{ i_k j_k } \mathbf{G}_{i_k j_k}^T$. While the polar decomposition achieves a symmetric positive definite result, our approach diagonalizes the matrix, i.e., $\lim_{k \to \infty} \mathbf{X}_k = \begin{bmatrix} \mathbf{\Sigma}_p & \mathbf{0}_{p \times (N-p)} \\ \mathbf{0}_{(n-p) \times p} & \mathbf{Y} \end{bmatrix}$ where $\mathbf{\Sigma}_p$ is a diagonal containing the largest $p$ singular values and $\mathbf{Y} \in \mathbb{R}^{(n-p) \times (N-p)}$ obeys $\|  \mathbf{Y} \|_F^2 = \sum_{k=p+1}^n \sigma_k^2$. We note that the work in \cite{rusu2020constructing, rusu2021iterative} deals with computing the eigenvalue decomposition for symmetric matrices, i.e., when $\mathbf{X}_k$ is square and symmetric, and we also have that $\mathbf{G}_{ i_k j_k } = \mathbf{H}_{ i_k j_k }^T$. $\hfill \blacksquare$
	
	Due to the factorization in \eqref{eq:bothfactorizations} we will see a trade-off between the sparsity of the singular vector and their accuracy (how well they diagonalize the data matrix). Lower $m$, say order $O(p)$ or $O(p\log_2 n)$, will lead to highly sparse singular vectors that provide a rough approximation while large $m$, say $O(np)$ or $O(p^2)$, will provide better approximation but fuller vectors(assuming that the true singular vectors are indeed full).
	
	We next describe the complete proposed algorithm.
	
	\subsection{The proposed algorithm}
	
	In this section, we describe the detailed proposed algorithm. The idea is to iteratively and approximately solve \eqref{eq:optimizationproblem} where the working variables (the singular vectors) are explicitly factored as in \eqref{eq:bothfactorizations} and each iteration performs the locally optimal step using Result 2. For simplicity, we assume the total number of basic transformations $m$ is given and fixed.
	
	\begin{algorithm}[t]
		\caption{\newline \textbf{Input: }Data matrix $\mathbf{X} \in \mathbb{R}^{n \times N}$, size of rank approximation $p \in \mathbb{N}^*$, and number of basic transformations/iterations $m \geq 1$.\newline \textbf{Output: } Approximate left singular space $\mathbf{\bar{U}} \in \mathbb{R}^{n \times n}$ as \eqref{eq:bothfactorizations}.}
		\begin{algorithmic}
			\State \textbf{1. }Initialize: $\mathbf{\bar{U}} \leftarrow \mathbf{I}_n$, $\mathbf{X}_{0} \leftarrow \mathbf{X}$, compute $\mathcal{C}_{ij}$ with \eqref{eq:themin} for $N \geq j > i \geq 1$ and $i \leq p$.
			
			\State \textbf{2. }Iterative process, for $k = 1,\dots,m$:
			\begin{itemize}
				\item Find $(i_k^\star, j_k^\star) = \underset{i, j}{\arg \max}\ \mathcal{C}_{ij}$.
				\item Get $\mathbf{G}_{i_k^\star j_k^\star}$ and $\mathbf{H}_{i_k^\star j_k^\star}$ according to Result 2 with $\mathbf{X}_{k-1}$.
				\item Update $\mathbf{X}_{(k)} \leftarrow \mathbf{G}_{i_k^\star j_k^\star}^T \mathbf{X}_{(k-1) } \mathbf{H}_{i_k^\star j_k^\star}$, $\mathbf{\bar{U}} \leftarrow \mathbf{\bar{U}}\mathbf{G}_{i_k^\star j_k^\star}$.
				\item With \eqref{eq:themin}, update $\mathcal{C}_{i_k^\star j}$ for $N \geq j > i_k^\star$, $\mathcal{C}_{ij_k^\star}$ for $i \leq p$.
			\end{itemize}
		\end{algorithmic}
	\end{algorithm}
	
	The complete proposed procedure is shown in Algorithm 1. From the output $\mathbf{\bar{U}}$ we keep the first $p$ columns, i.e., $\mathbf{\bar{U}}_p$. The initialization step takes $O(p N)$ operations due to the calculations of all the scores $\mathcal{C}_{ij}$ and we note that this step is trivially paralellizable. Then, each iteration takes $O(n+N)$: the updates of a subset of scores and the calculation of $\mathbf{X}_k$ from $\mathbf{X}_{k-1}$ are easy as only two rows/columns are updated at each step. The overall complexity of Algorithm 1 is therefore order $O(pN + m(n + N))$. If we choose $m \sim O(p \log_2 N)$ then the overall complexity is $O( pN + pn\log_2 N + pN \log_2 N )$. In terms of memory consumption, the $m$ transformations accummulate in $\mathbf{X}$ while for $\mathbf{\bar{U}}$ we store an extra $O(m)$ bits or $O(np)$ if stored explicitly, whichever is more convenient.
	
	\noindent \textbf{Remark 4 (Block transformations).} Algorithm 1 is designed to be numerically efficient for $m \sim O(p)$ or $m \sim O(p \log_2 N)$. In principle, the algorithm could also be used to compute the non-sparse approximate SVD of $\mathbf{X}$ but due to the $2 \times 2$ updates the number of iterations $m$ would be significant, for example $O(p N)$. In this case, a block version of the algorithm could be used. This approach, which we call Algorithm 1 - block, would pair each index $i \leq p$ with another index $j > p$ by choosing the next largest $\mathcal{C}_{ij}$ on new indices and thus perform a block SVD of size $2p \times 2p$ (here we could use a standard SVD algorithm). The benefit is that we exchange a single SVD of size $n \times N$ with a series of smaller sized SVDs $2p \ll n$.$\hfill \blacksquare$
	
	%---------------------------------------------------------------------
	\section{Experimental results}
	%---------------------------------------------------------------------
	\begin{figure*}[!h]
		\centering
		\includegraphics[trim = 75 10 60 22, clip, width=0.8\textwidth]{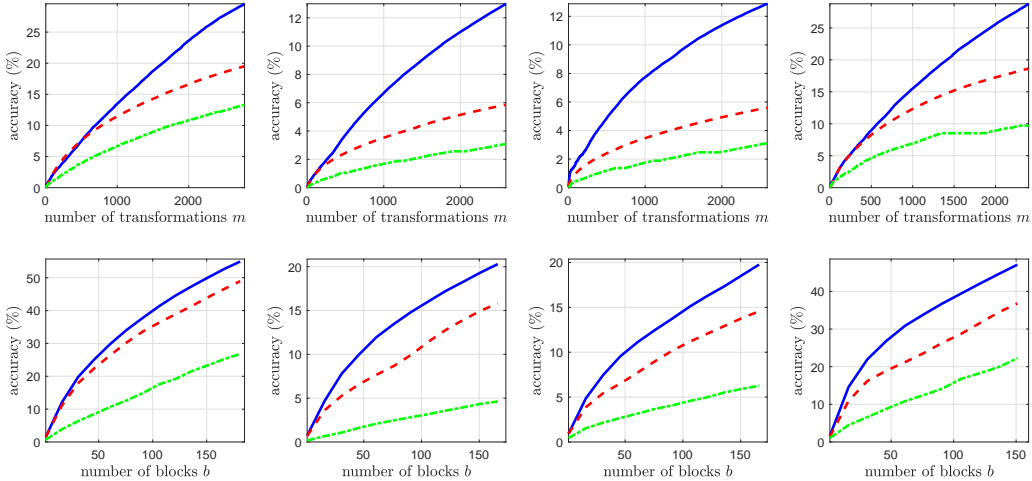}
		\caption{We show the accuracy, as defined in \eqref{eq:error}, achieved by the proposed method (blue solid) against the Kogbetliantz (red dashed) and randomized (green dashed-dot) methods. The accuracy improves with the number of transformations $m$ (top) or transformation blocks $b$ of size $2p\times 2p$ (bottom), see Remark 4. From left to right we have results for the ISOLET, MNIST digits, MNIST fashion, and USPS datasets, respectively. In all cases we have $p = 15$.}
		\label{fig:figure1}
	\end{figure*}
	
	In this section we provide experimental results using popular datasets from the machine learning community to highlight the performance of the proposed method\footnote{https://github.com/cristian-rusu-research/JACOBI-SVD}. We use the following datasets: ISOLET\footnote{https://www.archive.ics.uci.edu/ml/datasets/isolet} with 26 classes and $n = 617, N = 7797$; USPS\footnote{https://www.github.com/darshanbagul/USPS\_Digit\_Classification} with 10 classes and $n = 256, N = 9298$; EMNIST digits\footnote{https://www.nist.gov/itl/iad/image-group/emnist-dataset} with 10 classes and $n = 784, N = 6 \times 10^4$; and EMNIST fashion\footnote{https://www.github.com/zalandoresearch/fashion-mnist} with 10 classes and $n = 784, N = 6 \times 10^4$.

	To measure the performance of the proposed algorithm and competitors, we will use the following accuracy metric:
	\begin{equation}
		\epsilon = \frac{ \text{tr}( \mathbf{W}^T \mathbf{X}_m ) }{ \sum_{k=1}^p \sigma_i } (\%),
		\label{eq:error}
	\end{equation}
	the denominator represents the sum of the true largest $p$ singular values while the numerator is the sum of the first $p$ diagonal elements from $\mathbf{X}_m$, i.e., the data matrix $\mathbf{X}$ after $m$ steps of the proposed algorithm have been applied.
	
	In Figure \ref{fig:figure1} we show the evolution of the error \eqref{eq:error} with the number of transformations $m$ for both Algorithm 1 and Algorithm 1 - block. For comparisons and reference we show the Kogbetliantz method (indices $(i_k, j_k)$ are chosen to maximize $|X_{i_k j_k}^{(k)}| + |X_{j_k i_k}^{(k)}|$) and a randomized approach (indices $(i_k, j_k)$ are chosen uniformly at random), respectively. The proposed method performs best, while Kogbetliantz picks many times indices that lead to $\mathbf{G}_{i_k j_k} = \mathbf{I}_n$ (see Remark 1). Due to their prevalence in the numerical linear algebra literature \cite{10.1145/2842602}, we also test against a randomized method that picks indices $(i_k, j_k)$ uniformly at random and then performs the optimal transform on those indices. Similar results are observed irrespective of the choice for $p$.
	
\begin{figure}[!t]
		\centering
		\includegraphics[trim = 48 0 58 15, clip, width=0.5\textwidth]{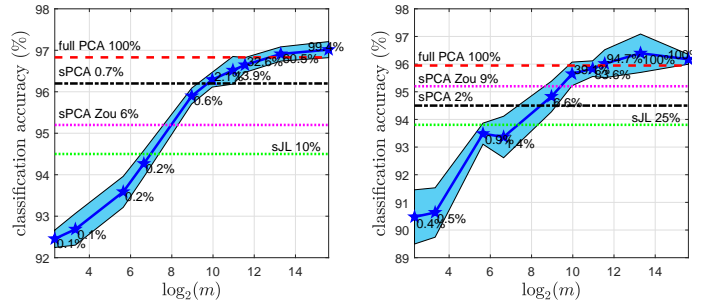}
		\caption{Average classification accuracy achieved by $k$-NN after dimensionality reduction is done with the proposed method choosing $p = 100$ and $p = 40$ for the MNIST (left) and USPS (right) datasets, respectively. Percentages show the fill-in of the projection.}
		\label{fig:figure2}
	\end{figure}

	In Figure \ref{fig:figure2} we show an application of the singular vectors we compute via the proposed method for dimensionality reduction with $p = 100$ in a classification scenario. We consider the USPS and MNIST digits datasets for which we use the $k$-Nearest Neighbors ($k$-NN) algorithm with $K = 25$ to peform classification. We randomly split the dataset into $N_\text{train} = 50000$ and $N_\text{test} = 10000$ for MNIST, and $N_\text{train} = 7291$ and $N_\text{test} = 2007$ for USPS. As performance reference, we show the results of the full PCA and the sparse JL (sJL) transforms \cite{DBLP:journals/corr/abs-1012-1577} with $s = 10$ non-zeroes per column. All results are averaged over 10 realizations. We also compare against the sparse PCA (sPCA) \cite{Hein2010AnIP} and one of the earlier approaches (sPCA Zou) \cite{doi:10.1198/106186006X113430} from the literature.
	
	It is surprizing how very sparse projections (below $1\%$ fill-in) lead to within a few percentages of the state-of-the-art classification results. Both $p$ and $K$ were chosen such that the full PCA and $k$-NN lead to the best experimental results.
	
	We report running times for the proposed method below one second when $\log_2(m) \leq 8$ while for the largest $m$, i.e., $\log_2(m) = 16$, we have 300 and 6 seconds for MNIST and USPS, respectively. sPCA takes 540 and 8 seconds while sPCA Zou takes 450 and 6 seconds for MNIST and USPS, respectively. Both previous methods are slower than the proposed method. We would also like to highlight that there are no hyperparameters to tune for Algorithm 1.
	
	%---------------------------------------------------------------------
	\section{Conclusions}
	%---------------------------------------------------------------------
	In this paper, we have proposed a new algorithm to find a few approximate sparse principal components from a data matrix which we have used to perform dimensionality reduction. The proposed method works directly on the data matrix and does not explicitly build the covariance matrix of the data, and is therefore memory efficient. Numerical results that build sparse principal components and then perform dimensionality reduction with application to classification show that the method is practical and compares very well against the state-of-the-art literature, especially in the running time.
	
	\newpage
	\bibliographystyle{IEEEbib}
	\bibliography{bib, refs}
	
\end{document}